\documentclass[11pt]{article}

\usepackage[final]{acl}

\usepackage{times}
\usepackage{latexsym}
\usepackage{makecell}
\usepackage[most]{tcolorbox}
\usepackage{xcolor}
\usepackage{placeins}
\usepackage{float}
\usepackage[T1]{fontenc}

\usepackage[utf8]{inputenc}

\usepackage{microtype}

\usepackage{inconsolata}

\usepackage{graphicx}
\usepackage{listings}
\usepackage{xcolor}

\lstdefinelanguage{json}{
    basicstyle=\ttfamily\small,
    numbers=left,
    numberstyle=\tiny,
    stepnumber=1,
    numbersep=5pt,
    showstringspaces=false,
    breaklines=true,
    frame=single,
    literate=
     *{0}{{{\color{blue}0}}}{1}
      {1}{{{\color{blue}1}}}{1}
      {2}{{{\color{blue}2}}}{1}
      {3}{{{\color{blue}3}}}{1}
      {4}{{{\color{blue}4}}}{1}
      {5}{{{\color{blue}5}}}{1}
      {6}{{{\color{blue}6}}}{1}
      {7}{{{\color{blue}7}}}{1}
      {8}{{{\color{blue}8}}}{1}
      {9}{{{\color{blue}9}}}{1}
      {:}{{{\color{red}{:}}}}{1}
      {,}{{{\color{red}{,}}}}{1}
      {"}{{{\color{gray}{"}}}}{1}
}

\title{psytechlab at CLPsych 2026: Utilising Natural Language Processing methods and Large Language Models for Social Media Text Analysis}

\author{Igor Buyanov \\
  Psytechlab \\
  \texttt{i.buyanov@psytechlab.co} \\\And
  Nafisa Valieva \\
  Psytechlab \\
  \texttt{n.valieva@psytechlab.co} \\\And
  Ekaterina Mazurina \\
  Psytechlab \\
  \texttt{e.mazurina@psytechlab.co}}

\begin{document}
\maketitle
\begin{abstract}
Social media posts are a rich and valuable source of data for analyzing mental health states and users' well-being using automated analysis tools. In this work, we demonstrate how we used a range of Natural Language Processing (NLP) methods, including Long Short-Term Memory (LSTM), BERT-based models, and Large Language Models (LLMs), for self-state and well-being analysis and summarization during the CLPsych Shared Task 2026. Our approach achieved one of the top Consistency and Contradiction scores for the summarization task and also middle-level results for the other tasks. By testing and developing such mental health-state estimation systems, we contributed to improving mental health support systems. We make our code available\footnote{\url{https://github.com/psytechlab/CLPsych2026/}}.
\end{abstract}

\section{Introduction}

In modern times, mental health problems are recognized as a major topic. The WHO statistics say that more than 1 billion people suffer from mental health disorders, which is 1 in 7 people globally \cite{WHOMental}. These statistics gain the researcher's attention from many scientific fields to address these issues. In particular, AI researchers are actively exploring ways to develop automated tools for diagnostic and therapeutic purposes to meet this demand.

To this day, researchers have achieved notable success in the diagnosis of various mental disorders, from depression and anxiety to bipolar disorder and schizophrenia. The main source of data is social media like Reddit\footnote{\url{https://www.reddit.com/}}, X\footnote{\url{https://x.com/}}, and other platforms. These platforms allow users to discuss many topics, including their mental states. This fact is leveraged by the researchers to bound certain users' textual productions to their self-disclosed mental states. Upon these datasets, the machine learning techniques can be applied to reveal various language patterns. Such models being trained can be used to identify users with mental health issues at scale, only by the data they produce, allowing for targeted help. Moreover, the growing power of LLMs can help researchers and practitioners achieve notable results in reasoning and analyzing data, offering insights into individuals' mental states from their textual data.

Unfortunately, the described approach also has drawbacks. The first one is potential label noise, as the researcher relies solely on user-provided information, which cannot be reliably confirmed. On the other hand, collecting clinically reliable annotated data raises the expenses and ethical challenges.

The second drawback is that researchers must associate all user data with a single static label. In reality, the mental state has a dynamic nature, and so many mental issues are better seen in their changes to understand their severity. Another side of this drawback is that not all content is actually meaningful from a diagnostic perspective, as social media platforms are primarily built for other purposes. 

Luckily, the CLPsych Shared Task 2026 \cite{ali026overview} provides researchers with the opportunity to work with the dataset without the aforementioned drawbacks. Specifically, the organizers suggest three challenging tasks: Adaptive and Maladaptive ABCD element Prediction, Moment of Change (MoC) identification, and Summary of changes in a timeline. The provided dataset consists of the user's post time series, professionally annotated using the MIND framework \cite{atzil_slonim_2025_mind}, a conceptual model that represents the human experience as a continually evolving series of self-states.

In this paper, we describe our system as psytechlab team for solving this task. Briefly, our system combines various approaches and models. By the end of the Shared Task, our approach beat almost all baselines, though it remains far from the best solution. All prompts that we used can be found in Appendix \ref{sec:appendix_prompts}.

\section{Methods}
\begin{table*}[]
\centering
\begin{tabular}{llll}
\hline
\textbf{Task} & \textbf{Metric} &  \textbf{Score}  & \textbf{Rank} \\ \hline
Task 1.1               & Average Subelement Macro F1       & 0.274  & 12 (17) \\ 
Task 1.2               & Avg RMSE Maladaptive + Adaptive   & 1.407  & 15 (17) \\ 
Task 2                 & Combined (Post/Timeline) Macro F1 & 0.372  & 15 (18) \\ 
Task 3.1               & Score Rank Average                & 7.3    & 8 (13)  \\ 
Task 3.2 Improvments   & Overall Score                     & 0.544 & 5 (9)   \\ 
Task 3.2 Deterioration & Overall Score                     & 0.491 & 6 (9)   \\ \hline
\end{tabular}
\caption{Overall results and ranks for CLPsych 2026 Shared Task. The overall rank range is shown in parentheses.}
\label{tab:my-table}
\end{table*}

\subsection{Task 1}

The goal of this task is twofold: to predict the dominant ABCD subelements and the self-state composition, and to estimate the presence value at scale 1-5, indicating the extent to which the self-states are expressed in the post. The ABCD stands for \begin{itemize}
    \item Affect (A): emotional tone or mood.
    \item Behavior(B), which is split into a toward Self (B-S) and Others (B-O): actions or tendencies directed inward or outward.
    \item Cognition (C), which is split into a toward Self (C-S) and Others (C-O): beliefs, interpretations, and appraisals.
    \item Desire (D): motivations, needs, wishes, and expectations.
\end{itemize}

\textbf{Metrics for Task 1}. The first task result is scored by many aspects; here, we provide only for ranking: (1) compute F1 per element, (2) Macro average across 6 elements within each valence, (3) Average of Adaptive and Maladaptive Macro F1. The second task is scored by MAE, RMSE, Quadratic Weighted Kappa (QWK), and Spearman correlation. The final ranking is the mean of the per-valence RMSEs.

\textbf{Submission 1}. We apply the pipeline approach for this task. First, we split the post into sentences. Then, each sentence is classified into one of the subelements or the irrelevant class. Finally, we aggregate the sentence predictions to determine the most dominant subelement using subelement-specific weights, obtained from the source dataset by calculating each subelement's presence proportion. Having a subelement, we can map it directly to the state and valence.

We train a classifier to predict all existing subelements. Because the training data was insufficient, we used two strategies to augment the dataset. The first is to translate the part of the dataset from \cite{dataset_meth_25}, where some classes are semantically aligned with subelements. The second one is to directly generate data using local Qwen3.5-35B-A3B \cite{qwen35-35b-a3b} with an 8-shot prompt using evidence as examples. We generate 500 texts for each class. Then we merge all data into a single dataset and split it into training and test sets. We experiment with BERT \cite{devlin-etal-2019-bert} and ModernBERT \cite{modernbert}, and they show comparative performance: 0.79 F1 macro for BERT and 0.80 F1 macro for ModernBERT. 

Speaking of aggregating schema, to obtain weights, we split all the training text into sentences, assign them a subelement if they overlap with the evidence, and annotate a piece of post text for a particular subelement. If the sentence doesn't overlap with any evidence, it's assigned as irrelevant. Next, we calculate the proportion of subelements grouped by valence and state. The irrelevant class is assigned a minimum value across all weights.

For inference, we obtain predictions for each sentence and sum the weights across all predicted subelements and the irrelevant class. Next, we take a maximum aggregation for the subelement weights grouped by valence and state. At this point, we have two dominant subelements  for each valence. Finally, we exclude subelements if their cumulative weight is below the irrelevant cumulative weight. 

To estimate the presence value for each valence, we use ModernBERT in regression mode and train it on available data. Thanks to ModernBERT's wide context length, we can use the entire post text. We draw inspiration from previous CLPsych Shared Task solutions \cite{Chakraborty2025SelfStateEE}, where this approach has proven promising.

\textbf{Submission 2}. Here we prompt the local Qwen3.5-35B-A3B to solve the task, providing it with the names of the subelements. We use the same ModernBERT model to estimate the valence presence.

\subsection{Task 2}
In Task 2, participants were given a chronologically ordered sequence of posts and asked to detect \textit{Switch} or \textit{Escalation} in each post in the timeline. 

\textbf{Metrics for Task2}. Switch and Escalation labels are evaluated as independent binary classification tasks. In post-level evaluation, precision, recall, and F1-score per label are pooled across all posts. For timeline-level evaluation, precision, recall, and F1-score per label are macro-averaged across timelines. 

\textbf{Submission 1}.
To obtain post representations, we split the post text into sentences and combine embeddings from MiniLM-L6-v2 \cite{reimers-2019-sentence-bert} and concatenate the probability vectors from several models as it was done in \cite{bayram-benhiba-2022-emotionally}: BERTweet Sentiment \cite{perez2021pysentimiento}, EmoRoBERTa \cite{ghoshal2022emoroberta}, and twitter-roberta-emotion \cite{barbieri-etal-2020-tweeteval}. Then, sentence vectors are averaged to obtain the final post text representation. Next, we use the BiLSTM \cite{lample-etal-2016-neural} model to predict Switch and Escalation separately. To obtain optimal hyperparameters, we use Optuna \cite{akiba2019optuna} for optimization.

\textbf{Submission 2}.
This is the same architecture as Submission 1. The difference is the special feature set, which includes temporal features of the posts.

\textbf{Additional submissions}.
Here, we primarily experiment with the Tempoformer model \cite{tempoformer}, with which we make three submissions. Another model we experimented with was HoRoBERT \cite{horobert}.

\subsection{Task 3}
In Task 3.1, the participants must generate a timeline-level summary describing patterns of self-state dynamics and their progression over time within a sequence of posts surrounding a change event (Switch/Escalation). The Task 3.2 goal was to identify and summarize recurrent dynamic patterns of deterioration or improvement that recur across multiple sequences.

\textbf{Metrics for Task3.1}.
\textit{Contradiction Score:} For each predicted sentence, the mean NLI (Natural Language Inference) contradiction probability is computed against all gold summary sentences, then the maximum is taken across predicted sentences and sequences. \textit{Consistency Score} is equal to $1 - \texttt{mean\_contradiction}$. \textit{Rouge-L Recall} \cite{lin-2004-rouge} score was computed to measure temporal word coverage and \textit{BERTScore Recall} \cite{zhang2020bertscore} for semantic content coverage.  

\textbf{Metrics for Task3.2}. The results for Task 3.2 are evaluated by humans using the following criteria. \textit{Fit of Evidence Support} evaluates how well the proposed signature is supported by the sequences submitted as evidence. \textit{Recurrence score} evaluates how recurrent the proposed signature is across sequences. \textit{Specificity score} evaluates how specific and non-generic the proposed signature is. The overall score for this task is computed using the formula $0.5 \cdot \texttt{Fit} + 0.5 \cdot \texttt{HarmonicMean(Recurrence, Specificity)}$.

\textbf{Submission 1}. We used zero-shot prompting to prompt Qwen3.5-35B-A3B model to summarize timeline of posts (no predictions from Task 1 and Task 2 were used), however, in the system instruction we provided MIND framework description. 

\textbf{Submission 2}. At first, we used predictions from Task 1 and a few-shot prompting to make an intermediate summary of each post in a timeline using the Qwen3.5-35B-A3B model. Then, given intermediate summaries of each post in a timeline, we also got a full summary of the timeline using few-shot prompting and the Qwen3.5-35B-A3B model.

\textbf{Submission 3}. We used the default prompt from \cite{teamblue} CLPsych 2025 shared tasks \cite{tseriotou-etal-2025-overview} best submission for timeline summarization. The model instruction lacks conceptual explanations or additional context to support a timeline-level summary. We also used the Llama-3.2-3B-Instruct \cite{llama32report} model for this task.

\textbf{Additional sumbissions}. The setting is the same as in Submission 3, but we used the Qwen3.5-35B-A3B model instead of Llama-3.2-3B-Instruct.

For Task 3.2, we used the Llama-3.2-3B-Instruct model. Firstly, LLM was instructed to detect the direction of well-being change in each timeline (deterioration, improvement, mixed, or neutral). After that, we instructed the LLM to detect ABCD elements (with no subelements) and assign a self-state label (adaptive or maladaptive). After that, we obtained statistics on the number of adaptive and maladaptive states present in each ABCD element type and each self-state ratio across well-being change types. Then we found CE (Change Event) index in a timeline based on the number of adaptive and maladaptive states in each post. Then, a summary of example timelines with improvement and deterioration change types was obtained using LLM.

\section{Results and analysis}

The overall results are shown in Table ~\ref{tab:my-table}. The per-task submission results are in Tables ~\ref{task11:submission1}, ~\ref{task12:submission1}, ~\ref{task2:submission1}, ~\ref{task3.1:submission1}, ~\ref{task3.2:submission1} in Appendix \ref{sec:appendix_results}. Our system outperforms all baselines except Tempoformer on Task 2, though its rank is at the lower intermediate level. Here, we conduct an analysis to identify potential reasons.

For Task 1.1, the main caveat is the long text, which can contain multiple subelements. We split the text into sentences and aggregate the results, but the classifier, despite a good overall metric, poorly recognizes the irrelevant class. As irrelevant sentences are far more than relevant ones, their poor recognition leads to heavy noise at the aggregation step. The reason is that the irrelevant class is not bounded by any rules, so the potential volume of this class is huge. If the irrelevant class doesn't have enough negative examples where patterns similar to the relevant classes occur, but they are only similar, the model will make false positive predictions on such texts. As evidence, the confusion matrix shows that the irrelevant class is confused with 23 out of 32 subelements, mostly from 1 to 5 percent, and 11 with the "Relating behaviour" subelement.

The bad result of Task 1.1 influences Task 1.2 because the presence value should be estimated for where the valence is predicted to be at all. Actually, on the validation set we form from the overall training data, ModernBERT shows 0.824 RMSE on Adaptive data and 0.904 RMSE on Maladaptive data.

The main weakness of Task 2 is the model, which cannot capture all necessary dependencies. In particular, the Switch performance is twice as poor as Escalation, which heavily undermines the overall metric. We hypothesize that it's due to label formulation. The Escalation label is defined as a gradual intensification of mood over a sequence of consecutive posts. Mood expressions can be directly bound to certain lexical patterns. On the other hand, the Switch label is defined as the difference in well-being scores between two consecutive posts exceeding 2. The well-being score is based on the Global Assessment of Functioning, which means there is an intermediate hidden layer between the label being predicted and the text. It seems the model lacks the power to identify the necessary dependencies between the text and the well-being difference.

Another issue that we faced was the Tempoformer training. We could not manage to properly train it during the competition time, but succeeded at the analysis time, as can be seen in Table ~\ref{task2:submission1}. Still, the organizer Tempoformer baseline is much higher than our results. This highlights the importance of a careful understanding of the training process.

Speaking of Task 3, we were surprised that a comparably small and old model with a simple prompt can achieve high results on metrics that show the form of a summary. The main consideration here is to find a way to incorporate the content data without losing the accurate form of the summary.

\section{Conclusion}

We have presented a mixed system that used various NLP techniques on each Shared Task. Across all tasks, our system outperformed almost all baselines and achieved notable performance on two metrics in Task 3.1. Nevertheless, there are many points to improve the solution that we highlight in the analysis section.

\section{Limitations}

This study has several limitations. First, although the shared task data are more structured than typical social media mental health corpora, the target labels still simplify complex and dynamic psychological processes. Thus, model outputs should be interpreted as approximations of annotated self-state patterns rather than as direct evidence of a person's mental condition.

Second, our methods are constrained by the limited amount of training data. In Task~1, sentence-level splitting and aggregation likely introduced additional noise, especially due to weak recognition of irrelevant sentences. In addition, the use of translated and synthetically generated examples may have introduced distributional mismatch. In Task~2, the comparatively weak results suggest that our temporal models did not fully capture longer-range dependencies and subtle changes across timelines. For Task~3, strong automatic summarization metrics do not guarantee clinically faithful summaries, since LLMs may omit important evidence or produce plausible but unsupported statements.

Third, this work was conducted in a restricted-access environment, which limited the range of models and experiments available to us. In particular, large language models were used only in locally deployed form; no external API-based or cloud-hosted LLMs were used. While this reduced potential data exposure, it also constrained experimentation.

\section{Ethical Considerations}

From an ethical perspective, this work deals with highly sensitive mental health-related text and should be treated with appropriate care. Even when shared for research, such data may contain intimate personal information. Using only local LLMs in a restricted-access environment was an important privacy-preserving measure in our study. However, privacy protection alone does not remove other risks. The models may reflect demographic, cultural, or linguistic biases present in the data and in pretrained models, leading to harmful false positives or false negatives.

Finally, our systems are intended strictly for research purposes and not for clinical diagnosis or autonomous decision-making. Predictions and generated summaries should not be treated as definitive assessments of an individual's mental health and should not replace qualified professional judgment.

\bibliography{custom}
\nocite{atzil_slonim_2026,tsakalidis-etal-2022-overview,atzil_slonim_2025_mind,tseriotou-etal-2025-overview,ali026overview}

\clearpage
\onecolumn
\appendix
\section{Appendix. Detailed CLPsych 2026 Shared Task Results}
\label{sec:appendix_results}

\begin{table}[H]
  \centering
  \begin{tabular}{lllll}
    \hline
    \textbf{Team} & \textbf{\makecell{Average \\ Subelement \\ Macro F1}} &  \textbf{\makecell{Adaptive \\ Subelement \\ Macro F1}}  & \textbf{\makecell{Maladaptive \\ Subelement \\ Macro F1}} & \textbf{Rank} \\
    \hline
    Submission 1     & 0.274 & 0.263&0.285&12         \\
    Submission 2     & 0.139 & 0.139  &0.140&-     \\
    CUNY (Best) & 0.442  & 0.388 &0.496&1 \\
   \hline
  \end{tabular}
  \caption{Task 1.1 submission results and comparison with the best performing solutions}
  \label{task11:submission1}
\end{table}

\begin{table}[H]
  \centering
  \begin{tabular}{lllll}
    \hline
    \textbf{Team} & \textbf{\makecell{Average \\ RMSE}} &  \textbf{\makecell{Adaptive \\ RMSE}}  & \textbf{\makecell{Maladaptive \\ RMSE}} & \textbf{Rank} \\
    \hline
    Submission 1     & 1.407 & 1.389&1.424&15         \\
    Submission 2     & 1.416 & 1.379  &1.45&-     \\
    Meronym Labs (Best) & 0.917  & 0.833 &1&1 \\
   \hline
  \end{tabular}
  \caption{Task 1.2 submission results and comparison with the best performing solutions}
  \label{task12:submission1}
\end{table}

\begin{table}[H]
  \centering
  \begin{tabular}{lllll}
    \hline
    \textbf{Team} & \textbf{\makecell{Combined \\ Macro F1} } &  \textbf{\makecell{Post \\ Macro F1} }  & \textbf{\makecell{Timeline \\ Macro F1}} & \textbf{Rank} \\
    \hline
    Submission 1     & 0.372 & 0.371&0.372&15         \\
    Submission 2     & 0.324 & 0.316  &0.332&-     \\
    Submission 3 (tempoformer)    & 0.215  & 0.197 &0.233&-  \\
    Submission 4 (HoRoBERT)     & 0.405  & 0.464 &0.346&- \\
    Submission 5 (tempoformer)    & 0.414  & 0.439 &0.389&-  \\
    Submission 6 (tempoformer)    & 0.429  & 0.498 &0.36&-  \\
    USAI (Best) & 0.6  & 0.639 &0.561&1 \\
   \hline
  \end{tabular}
  \caption{Task 2 submission results and comparison with the best performing solutions}
  \label{task2:submission1}
\end{table}

\begin{table}[H]
  \centering
  \begin{tabular}{lllllll}
    \hline
    \textbf{Team} & \textbf{CS} & \textbf{CT} & \textbf{\makecell{Rouge-L \\ Recall}} & \textbf{\makecell{BERTScore \\ Recall}} & \textbf{\makecell{Score \\ Average}} & \textbf{Rank} \\
    \hline
    Submission 1     & 0.692 & 0.819&0.220&0.279&0.343&-          \\
    Submission 2     & 0.659 & 0.885  &0.235&0.303&0.328&-      \\
    Submission 3     & 0.857  & 0.571 &0.078&0.147&0.378&8  \\
    Submission 4     & 0.768  & 0.808 &0.119&-&-&- \\
    MERONYM\_LABS (Best) & 0.801  & 0.659 &0.266&0.345&0.438&1 \\
   \hline
  \end{tabular}
  \caption{Task 3.1 submission results and comparison with the best performing solutions}
  \label{task3.1:submission1}
\end{table}

\begin{table}[H]
  \centering
  \begin{tabular}{lllllll}
    \hline
    \textbf{Team} & \textbf{Change Type} & \textbf{Rank} & \textbf{Fit score} & \textbf{Recurrence} & \textbf{Specificity} & \textbf{Overall} \\
    \hline
    Ours     & Improvement & 5&0.6875&1&0.25&0.5437          \\
    DreamerNLplus     & Improvement & 1  &0.625&0.8125&1&0.7608      \\
    Ours     & Deterioration  & 6 &0.625&0.625&0.25&0.4911  \\
    CSE\_IIT\_Ropar     & Deterioration  & 1 &0.875&0.5625&0.9375&0.7891 \\
   \hline
  \end{tabular}
  \caption{Task 3.2 submission results and comparison with the best performing solutions}
  \label{task3.2:submission1}
\end{table}

\FloatBarrier

\section{Appendix. Prompts}
\label{sec:appendix_prompts}
\begin{tcolorbox}[
  breakable,
  colback=gray!10,
  colframe=black,
  boxrule=0.5pt,
  arc=2pt,
  title=Task 1.1 Submission 1 prompt
]
\scriptsize

You are an experienced psychologist specializing in assessing the emotional tone of texts. You need to classify texts. Each text must be assigned most appropriate adaptive subelement and most appropriate maladaptive subelement. Description of the subelements with examples for classification:

Adaptive subelements: 

1.  Calm/ laid back

2.  Sad, Emotional pain, grieving

3.  Content, happy, joy, hopeful 

4.  Vigor / energetic

5.  Justifiable anger/ assertive anger, justifiable outrage

6.  Proud

7.  Feel loved, belong 

8.  Relating behavior

9.  Autonomous or adaptive control behavior 

10.  Self care and improvement 

11.  Perception of the other as related

12.  Perception of the other as facilitating autonomy needs 

13.  Self-acceptance and compassion 

14.  Relatedness

15.  Autonomy and adaptive control

16.  Competence, self esteem, self-care

Maladaptive subelements:

1.  Anxious/ fearful/ tense

2.  Depressed, despair, hopeless 

3.  Mania

4.  Apathic, don’t care, blunted 

5.  Angry (aggression), disgust, contempt

6.  Ashamed, guilty

7.  Feel lonely 

8.  Fight or flight behavior

9.  Over controlled or controlling behavior 

10.  Self harm, neglect and avoidance 

11.  Perception of the other as detached or over-attached 

12.  Perception of the other as blocking autonomy needs 

13.  Self criticism 

14.  Expectation that relatedness needs will not be met

15.  Expectation that autonomy needs will not be met 

16.  Expectation that competence needs will not be met

Your answer should contain only the names of the subelements; no additions, clarifications, explanations or other words are needed. 

Return ONLY a JSON object in this exact format:
\texttt{\{\{"adaptive": ["subelement"], "maladaptive": ["subelement"]\}\}}

Text for classification: \texttt{\{text\}}

\end{tcolorbox}


\begin{tcolorbox}[
  breakable,
  colback=gray!10,
  colframe=black,
  boxrule=0.5pt,
  arc=2pt,
  title=Task 3.1 Submission 1 prompt
]
\scriptsize

\#\#\# Task:

Your task is to summarize self-states for each timeline, given the sequence of posts on the timeline. Specifically, generate a summary focusing on the interplay between adaptive and maladaptive self-states along the timeline. Emphasize temporal dynamics focusing on concepts such as flexibility, rigidity, improvement, and deterioration. If applicable, describe the extent to which the dominance of the self-states changes over time and how changes in aspects (Affect, Behavior, Cognition, and Desire) contribute to these transitions. 

---

\#\#\# Definitions:

Self-states constitute identifiable units characterized by specific combinations of Affect, Behavior, Cognition, and Desire/Need (ABCD dimensions) that tend to be coactivated in a meaningful manner for limited periods of time.

- An adaptive self-state pertains to aspects of Affect, Behaviour, and Cognition towards the self or others, which is conducive to the fulfillment of basic desires/needs (D), such as relatedness, autonomy and competence.

- A maladaptive self-state pertains to aspects of Affect, Behaviour, and Cognition towards the self or others, that hinder the fulfillment of basic desires/needs (D).

\#\#\# ABCD dimensions:

1. Affect (A): The type of emotion expressed by the person.

- Adaptive Examples: Calm/Laid back, Emotional Pain/Grieving, Content/Happy, Vigor/Energetic, Justifiable, Anger/Assertive Anger, Proud.

- Maladaptive Examples: Anxious/Tense/Fearful, Depressed/Despair/Hopeless, Mania, Apathetic/Don’t care/Blunted, Angry (Aggressive, Disgust, Contempt),
Ashamed/Guilty.

2. Behavior of the self with the Other (BO) : The person’s main behavior(s) toward the other

- Adaptive Examples: Relating behavior, Autonomous behavior

- Maladaptive Examples: Fight or fight behavior, Overcontrolled/controlling behavior

3. Behavior toward the Self (BS): The person’s main behavior(s) toward the self

- Adaptive Examples: Self-care behavior

- Maladaptive Examples: Self-harm, Neglect, Avoidance behavior

4. Cognition of the Other (CO): The person’s main perceptions of the other

- Adaptive Examples: Perception of the other as related, Perception of the other as facilitating autonomy needs

- Maladaptive Examples: Perception of the other as detached or over attached, Perception of the other as blocking autonomy needs

5. Cognition of the Self (CS): How the person perceives themselves.

- Adaptive Examples: Self-acceptance and self-compassion

- Maladaptive Examples: Self-criticism

6. Desire (D): The person’s main desire, need, intention, fear or expectation

- Adaptive Examples: Relatedness, Autonomy and adaptive control, Competence, Self-esteem, Self-care

- Maladaptive Examples: Expectation that relatedness need will not be met, Expectation that autonomy needs will not be met, Expectation that competence needs will not be met

---

\#\#\# Guidelines for Output:

- Response Section: Provide an answer under the headings ‘\#\#\# Timeline Summary:‘.

- Format the answer as a single paragraph, making it clear and consise.

- The summary should be no more than 350 words. 

- Ensure the timeline summary captures the main points without unnecessary details.

- Ensure to explicitly state the ABCD dimensions when pointing out self-states

---

\#\#\# Example:

\#\#\#\#\# Input text:

1. \texttt{<post\_1\_in\_timeline>}

2. \texttt{<post\_2\_in\_timeline>}

...

n. \texttt{<post\_n\_in\_timeline>}

\#\#\#\#\# Output text:

\#\#\# Timeline Summary: \texttt{<example\_timeline\_summary>}

---

\#\#\# Analyze the following input text based on the given criteria.

\#\#\# Input Text:

\texttt{<posts>}

\#\#\# Output Text:

\end{tcolorbox}

\begin{tcolorbox}[
  breakable,
  colback=gray!10,
  colframe=black,
  boxrule=0.5pt,
  arc=2pt,
  title=Task 3.1 Submission 2 Post Summary Intermediate prompt
]
\scriptsize

You are a mental health expert and analyzing a patient’s social media post to determine their well-being, their dominant self-state
of either adaptive or maladaptive. Your task is to summarize self-states for the social media post below. Specifically, generate a summary of the interplay between adaptive and maladaptive states identified in the post. Begin by determining which self-state is dominant (adaptive/maladaptive) and describe it first. For each self-state, identify the central organizing aspect (A, B, C, or D) that drives the state and structure the summary around it. Describe how this central aspect influences the rest, emphasizing potential causal relationships between them. Then, proceed to the second self-state and follow the same approach. If the post contains only one self-state (either adaptive or maladaptive), summarize only that state. Note that the summary does not need to explicitly highlight A, B, C, or D, but should aim to naturally integrate these elements into the description. You have 2 analysis examples. 

---

\#\#\# Definitions:

Self-states constitute identifiable units characterized by specific combinations of Affect, Behavior, Cognition, and Desire/Need (ABCD dimensions) that tend to be coactivated in a meaningful manner for limited periods of time.

- An adaptive self-state pertains to aspects of Affect, Behaviour, and Cognition towards the self or others, which is conducive to the fulfillment of basic desires/needs (D), such as relatedness, autonomy and competence.

- A maladaptive self-state pertains to aspects of Affect, Behaviour, and Cognition towards the self or others, that hinder the fulfillment of basic desires/needs (D).

\#\#\# ABCD dimensions:

1. Affect (A): The type of emotion expressed by the person.

- Adaptive Examples: Calm/Laid back, Emotional Pain/Grieving, Content/Happy, Vigor/Energetic, Justifiable, Anger/Assertive Anger, Proud.

- Maladaptive Examples: Anxious/Tense/Fearful, Depressed/Despair/Hopeless, Mania, Apathetic/Don’t care/Blunted, Angry (Aggressive, Disgust, Contempt),
Ashamed/Guilty.

2. Behavior of the self with the Other (BO) : The person’s main behavior(s) toward the other

- Adaptive Examples: Relating behavior, Autonomous behavior

- Maladaptive Examples: Fight or fight behavior, Overcontrolled/controlling behavior

3. Behavior toward the Self (BS): The person’s main behavior(s) toward the self

- Adaptive Examples: Self-care behavior

- Maladaptive Examples: Self-harm, Neglect, Avoidance behavior

4. Cognition of the Other (CO): The person’s main perceptions of the other

- Adaptive Examples: Perception of the other as related, Perception of the other as facilitating autonomy needs

- Maladaptive Examples: Perception of the other as detached or over attached, Perception of the other as blocking autonomy needs

5. Cognition of the Self (CS): How the person perceives themselves.

- Adaptive Examples: Self-acceptance and self-compassion

- Maladaptive Examples: Self-criticism

6. Desire (D): The person’s main desire, need, intention, fear or expectation

- Adaptive Examples: Relatedness, Autonomy and adaptive control, Competence, Self-esteem, Self-care

- Maladaptive Examples: Expectation that relatedness need will not be met, Expectation that autonomy needs will not be met, Expectation that competence needs will not be met

Presence refers to the extent to which the self state influences the person’s expressed experience in the post, how strongly it was emphasized in the post. It captures how central the self state is to the overall experience described in the post. Presence ratings are independent. Presence reflects psychological centrality and experiential influence, not mere frequency of words. Both adaptive and maladaptive self states may receive high presence scores within the same post, if both strongly shape the expressed experience. If a self state is not expressed in the post, it receives a
score of 1.

---

\#\#\# Analysis 1:
\texttt{<post\_1>}

**Adaptive post segments:**

\texttt{<adaptive\_states\_presence1>}

\texttt{<adaptive\_element1>}

**Maladaptive post segments:**

\texttt{<maladaptive\_states\_presence1>}

\texttt{<maladaptive\_elements1>}

Summary:

\texttt{<example\_summary1>}

---

\#\#\# Analysis 2:

\texttt{<post\_2>}

**Adaptive post segments:**

\texttt{<adaptive\_states\_presence2>}

\texttt{<adaptive\_element2>}

**Maladaptive post segments:**

\texttt{<maladaptive\_states\_presence2>}

\texttt{<maladaptive\_elements2>}

Summary: 

\texttt{<example\_summary2>}

---

\#\#\# Important notes:

- Keep your analysis compact, but still informative. Analyze the post as a whole.

- Return ONLY the summary, nothing more.

- Do not add stylistic features, such as making a bold text.

- Before you create a final summary, make sure you understand the examples.

Now analyze the following patent post.

\text{<patient\_post>}

Adaptive post segments:

\texttt{<adaptive\_segments>}

Maladaptive post segments:

\texttt{<maladaptive\_segments>}

Summary:

\texttt{<fill your assessment here>}

\end{tcolorbox}

\begin{tcolorbox}[
  breakable,
  colback=gray!10,
  colframe=black,
  boxrule=0.5pt,
  arc=2pt,
  title=Task 3.1 Submission 2 Timeline Summary prompt
]
\scriptsize

\#\#\# Task:

Your task is to summarize self-states for each timeline, given the sequence of posts'analysis on the timeline. Specifically, generate a summary focusing on the interplay between adaptive and maladaptive self-states along the timeline. Emphasize temporal dynamics focusing on concepts such as flexibility, rigidity, improvement, and deterioration. If applicable, describe the extent to which the dominance of the self-states changes over time and how changes in aspects (Affect, Behavior, Cognition, and Desire) contribute to these transitions. 

—--

\#\#\# Definitions:

Self-states constitute identifiable units characterized by specific combinations of Affect, Behavior, Cognition, and Desire/Need (ABCD dimensions) that tend to be coactivated in a meaningful manner for limited periods of time.

- An adaptive self-state pertains to aspects of Affect, Behaviour, and Cognition towards the self or others, which is conducive to the fulfillment of basic desires/needs (D), such as relatedness, autonomy and competence.

- A maladaptive self-state pertains to aspects of Affect, Behaviour, and Cognition towards the self or others, that hinder the fulfillment of basic desires/needs (D).

\#\#\# ABCD dimensions:

1. Affect (A): The type of emotion expressed by the person.

- Adaptive Examples: Calm/Laid back, Emotional Pain/Grieving, Content/Happy, Vigor/Energetic, Justifiable, Anger/Assertive Anger, Proud.

- Maladaptive Examples: Anxious/Tense/Fearful, Depressed/Despair/Hopeless, Mania, Apathetic/Don’t care/Blunted, Angry (Aggressive, Disgust, Contempt),
Ashamed/Guilty.

2. Behavior of the self with the Other (BO) : The person’s main behavior(s) toward the other

- Adaptive Examples: Relating behavior, Autonomous behavior

- Maladaptive Examples: Fight or fight behavior, Overcontrolled/controlling behavior

3. Behavior toward the Self (BS): The person’s main behavior(s) toward the self

- Adaptive Examples: Self-care behavior

- Maladaptive Examples: Self-harm, Neglect, Avoidance behavior

4. Cognition of the Other (CO): The person’s main perceptions of the other

- Adaptive Examples: Perception of the other as related, Perception of the other as facilitating autonomy needs

- Maladaptive Examples: Perception of the other as detached or over attached, Perception of the other as blocking autonomy needs

5. Cognition of the Self (CS): How the person perceives themselves.

- Adaptive Examples: Self-acceptance and self-compassion

- Maladaptive Examples: Self-criticism

6. Desire (D): The person’s main desire, need, intention, fear or expectation

- Adaptive Examples: Relatedness, Autonomy and adaptive control, Competence, Self-esteem, Self-care

- Maladaptive Examples: Expectation that relatedness need will not be met, Expectation that autonomy needs will not be met, Expectation that competence needs will not be met

Presence refers to the extent to which the self state influences the person’s expressed experience in the post, how strongly it was emphasized in the post. It captures how central the self state is to the overall experience described in the post. Presence ratings are independent. Presence reflects psychological centrality and experiential influence, not mere frequency of words. Both adaptive and maladaptive self states may receive high presence scores within the same post, if both strongly shape the expressed experience. If a self state is not expressed in the post, it receives a
score of 1.

—--

\#\#\# Guidelines for Output:

- Response Section: Provide an answer under the headings ‘\#\#\# Timeline Summary:‘.

- Format the answer as a single paragraph, making it clear and consise.

- The summary should be no more than 350 words. 

- Ensure the timeline summary captures the main points without unnecessary details.

- Ensure to explicitly state the ABCD dimensions when pointing out self-states

—--

\#\#\# Example:

\#\#\#\#\# Input text:

1. \texttt{<intermediate\_post1\_summary>}

2. \texttt{<intermediate\_post2\_summary>}

...

N. \texttt{<intermediate\_postN\_summary>}

\#\#\#\#\# Output text:

\#\#\# Timeline Summary:

\texttt{<example\_timeline\_summary>}

—--

\#\#\# Analyze the following input text based on the given criteria.

\#\#\# Input Text:

\texttt{<intermediate\_posts\_summaries>}

\#\#\# Output Text:
\end{tcolorbox}

\begin{tcolorbox}[
  breakable,
  colback=gray!10,
  colframe=black,
  boxrule=0.5pt,
  arc=2pt,
  title=Task 3.1 Submission 3 prompt
]
\scriptsize

Given the following series of Reddit posts from one user, generate a timeline-level summary. Begin by determining which self-state is dominant (adaptive/maladaptive) and describe it first.

Timeline:

\texttt{<timeline\_text>}

Response format - ONLY json: \texttt{\{\{ "summary": "<timeline-level summary>" \}\}}
\end{tcolorbox}

\begin{tcolorbox}[
  breakable,
  colback=gray!10,
  colframe=black,
  boxrule=0.5pt,
  arc=2pt,
  title=Task 3.2 Well-being change direction detection prompt
]
\scriptsize

You are an expert clinical psychology annotator.

You classify summaries of user post-sequences into one of four categories describing the dominant direction of well-being change in the sequence.                                                        

Categories:

- deterioration: well-being decreases; maladaptive states gain strength; hopelessness, self-criticism, isolation, or symptom escalation emerge.

- improvement: well-being increases; adaptive states gain strength; hope, connection, coping, insight, or recovery emerge.

- mixed: both directions are clearly present (e.g., partial recovery followed by relapse, or vice versa) and neither dominates.

- neutral: no clear direction of change is described.

Return ONLY a JSON object with keys: 

  "change\_type": one of deterioration | improvement | mixed | neutral 
  
  "confidence": float in [0, 1]

  "rationale": one short sentence citing concrete elements from the summary

Summary:

\texttt{{summary\_text}}
\end{tcolorbox}

\begin{tcolorbox}[
  breakable,
  colback=gray!10,
  colframe=black,
  boxrule=0.5pt,
  arc=2pt,
  title=Task 3.2 ABCD elements and self-states detection prompt
]
\scriptsize

You are an expert clinical psychology annotator using the MIND (ABCD) framework with self-state analysis.

For each social media post, identify elements belonging to these categories:

- A (Affect): emotions, feelings, mood states

- B (Behavior): actions taken or avoided, including social actions

- C (Cognition): thoughts, beliefs, interpretations, self-talk 

- D (Desire/motivation): wants, intentions, goals, drives                                                                                 For each element, assign a self-state label:

- maladaptive: expresses hopelessness, self-criticism, isolation, avoidance, symptom escalation, rumination, worthlessness, risk 

- adaptive: expresses hope, connection, healthy coping, help-seeking, insight, agency, self-compassion, growth

One post can contain multiple elements of the same category, and can contain both adaptive and maladaptive elements simultaneously (this is called a dialogue between self-states -- mark both).

If a post has no codable ABCD content (e.g., a meme link, a factual question, pure small talk), return an empty "elements" list.

Return ONLY a JSON object:

\begin{lstlisting}[language=json]
{                    
    "elements": [
      {
       "type": "A"|"B"|"C"|"D", 
       "text": "short quote or paraphrase from the post",
       "state": "adaptive"|"maladaptive", 
       "confidence": 0.0-1.0
      }
    ]                            
}
\end{lstlisting}

\end{tcolorbox}

\end{document}